\documentclass[sigconf,nonacm]{acmart}
\AtBeginDocument{%
  }

\renewcommand\footnotetextcopyrightpermission[1]{}
\setcopyright{none}
\acmDOI{}
\acmISBN{}
\acmPrice{}

\usepackage[utf8]{inputenc} % allow utf-8 input
\usepackage[T1]{fontenc}    % use 8-bit T1 fonts
\usepackage{hyperref}       % hyperlinks
\usepackage{url}            % simple URL typesetting
\usepackage{booktabs}       % professional-quality tables
\usepackage{amsfonts}       % blackboard math symbols
\usepackage{nicefrac}       % compact symbols for 1/2, etc.
\usepackage{microtype}      % microtypography
\usepackage{booktabs}       % For professional table lines
\usepackage{multirow}       % For spanning multiple rows (optional)
\usepackage{siunitx}        % For alignment of numbers in columns (optional)
\usepackage[utf8]{inputenc} % allow utf-8 input
\usepackage[T1]{fontenc}    % use 8-bit T1 fonts
\usepackage{hyperref}       % hyperlinks

\usepackage{url}            % simple URL typesetting
\usepackage{booktabs}       % professional-quality tables
\usepackage{amsfonts}       % blackboard math symbols
\usepackage{nicefrac}       % compact symbols for 1/2, etc.
\usepackage{microtype}      % microtypography

\usepackage{amsmath,amsfonts,amsthm}
\usepackage{pifont}% http://ctan.org/pkg/pifont
\usepackage{soul}           % Strikethrough
\usepackage{booktabs}
\usepackage{tabularx}
\usepackage{multirow}
\usepackage{graphicx}

\usepackage{color}
\usepackage{graphicx}
\usepackage{subcaption}
\usepackage{makecell}
\usepackage{algorithm,bbm}
\usepackage{algorithmic}
\usepackage{multirow}
\usepackage{lineno}
\usepackage{enumerate}
\usepackage{placeins}

\DeclareMathOperator{\QL}{QL}

\def\benu{\begin{enumerate}}
\def\eenu{\end{enumerate}}

\newcommand\btheta{\boldsymbol \theta}
\begin{document}

%% The "title" command has an optional parameter,
%% allowing the author to define a "short title" to be used in page headers.
\title{SPADE-S: A Sparsity-Robust Foundational Forecaster}

%% The "author" command and its associated commands are used to define
%% the authors and their affiliations.
%% Of note is the shared affiliation of the first two authors, and the
%% "authornote" and "authornotemark" commands
%% used to denote shared contribution to the research.
\author{Malcolm Wolff\ }
\authornote{Authors contributed equally to this research.}
\authornote{Amazon SCOT Forecasting. New York, New York. USA.}
\email{wolfmalc@amazon.com}
\affiliation{%
  \institution{}
  \department{\vspace{-.05in}}
  \city{}
  \country{}
}
\author{Matthew Li}
\authornotemark[1]
\authornotemark[2]
\email{yumattli@amazon.com}
\affiliation{%
  \institution{}
  \department{\vspace{-.05in}}
  \city{}
  \country{}
}

\author{Ravi Kiran Selvam}
\authornotemark[1]
\authornotemark[2]
\email{ravisel@amazon.com}
\affiliation{%
  \institution{}
  \department{\vspace{-.05in}}
  \city{}
  \country{}
}

\author{Hanjing Zhu}
\authornotemark[1]
\authornotemark[2]
\email{hzhuad@amazon.com}
\affiliation{%
  \institution{}
  \department{\vspace{-.05in}}
  \city{}
  \country{}
}

\author{Kin G. Olivares}
\authornotemark[2]
\email{kigutie@amazon.com}
\affiliation{%
  \institution{}
  \department{\vspace{-.05in}}
  \city{}
  \country{}
}

\author{Ruijun Ma}
\authornotemark[2]
\email{ruijunma@amazon.com}
\affiliation{%
  \institution{}
  \department{\vspace{-.05in}}
  \city{}
  \country{}
}

\author{Abhinav Katoch}
\authornotemark[2]
\email{abkatoch@amazon.com}
\affiliation{%
  \institution{}
  \department{\vspace{-.05in}}
  \city{}
  \country{}
}

\author{Shankar Ramasubramanian}
\authornotemark[2]
\email{sramasub@amazon.com}
\affiliation{%
  \institution{}
  \department{\vspace{-.05in}}
  \city{}
  \country{}
}

\author{Mengfei Cao}
\authornotemark[2]
\email{mfcao@amazon.com}
\affiliation{%
  \institution{}
  \department{\vspace{-.05in}}
  \city{}
  \country{}
}

\author{Roberto Bandarra}
\authornotemark[2]
\email{rbmp@amazon.com}
\affiliation{%
  \institution{}
  \department{\vspace{-.05in}}
  \city{}
  \country{}
}

\author{Rahul Gopalsamy}
\authornotemark[2]
\email{rahulgo@amazon.com}
\affiliation{%
  \institution{}
  \department{\vspace{-.05in}}
  \city{}
  \country{}
}

\author{Stefania La Vattiata}
\email{steflvs@gmail.com}
\affiliation{%
  \institution{}
  \department{\vspace{-.05in}}
  \city{}
  \country{}
}

\author{Sitan Yang}
\authornotemark[2]
\email{sitanyan@amazon.com}
\affiliation{%
  \institution{}
  \department{\vspace{-.05in}}
  \city{}
  \country{}
}

\author{Michael W. Mahoney}
\authornotemark[2]
\email{zmahmich@amazon.com}
\affiliation{%
  \institution{}
  \department{\vspace{-.05in}}
  \city{}
  \country{}
}

% \affiliation{%
%   \institution{Institute for Clarity in Documentation}
%   \city{Dublin}
%   \state{Ohio}
%   \country{USA}
% }

% \author{Lars Th{\o}rv{\"a}ld}
% \affiliation{%
%   \institution{The Th{\o}rv{\"a}ld Group}
%   \city{Hekla}
%   \country{Iceland}}
% \email{larst@affiliation.org}

%%
%% By default, the full list of authors will be used in the page
%% headers. Often, this list is too long, and will overlap
%% other information printed in the page headers. This command allows
%% the author to define a more concise list
%% of authors' names for this purpose.
% \renewcommand{\shortauthors}{Trovato et al.}

%%
%% The abstract is a short summary of the work to be presented in the
%% article.
\begin{abstract}
  Despite significant advancements in time series forecasting, accurate modeling 
  of time series with strong %across high series-level 
  heterogeneity in 
  magnitude and/or sparsity %both magnitude and sparsity 
  patterns remains challenging for state of the art deep learning architectures.
  %
  % In this paper, we introduce \texttt{SPADE-S}, a robust forecasting architecture to explicitly address this shortcoming.
  %
  We identify several factors that lead existing models to systematically under-perform on low magnitude and sparse time series, including loss functions with implicit biases toward high-magnitude series, training-time sampling methods, and limitations of time series encoding methods.
  %
  % Based on our learnings, we develop a novel multi-head convolutional encoder and a model arm specifically designed to handle sparse multi-variate time series.
  To address these limitations, we introduce \texttt{SPADE-S}, a robust forecasting architecture with a novel multi-head convolutional encoder and a model arm specifically designed to handle sparse multi-variate time series.
  \texttt{SPADE-S} significantly reduces magnitude and sparsity-based systematic biases and improves overall prediction accuracy;
   empirical results demonstrate that \texttt{SPADE-S} outperforms existing state-of-the-art approaches across a diverse set of use-cases in demand forecasting.
  In particular, we show that, depending on the quantile forecast and magnitude of the series, \texttt{SPADE-S} can improve forecast accuracy by up to 15\%. This results in P90 overall forecast accuracy gains of 2.21\%, 6.58\%, and 4.28\%, and P50 forecast accuracy gains of 0.92\%, 0.77\%, and 1.95\% respectfully, for each of three distinct datasets, ranging from 3 million to 700 million series, from a large online retailer.
\end{abstract}

\maketitle

\section{Introduction}
%%%
% Deep Learning Models
%%%
State-of-the-art (SOTA) time series forecasting architectures have become capable of accurately modeling multivariate time series trajectories at scale in modern supply chain optimization applications.
Most notably:
Wen et al. \cite{wen2017mqrcnn} introduced \texttt{MQCNN}, a convolutional neural network-based architecture designed for probabilistic forecasting of multi-variate time series, demonstrating effectiveness over prior methods in capturing complex inter-dependencies among input co-variates and the target variable;
Eisenach et al. \cite{eisenach2020mqtransformer} proposed \texttt{MQTransformer}, which incorporates attention mechanisms to reduce forecast volatility and improve accuracy;
and Wolff et al. \cite{wolff2024spade} introduced \texttt{SPADE} (\texttt{S}plit \texttt{P}eak \texttt{A}ttention \texttt{DE}composition), an architecture which leverages exogenous future information and self-attention to separately address peak and off-peak demand dynamics.
\texttt{SPADE}, in particular, has shown ability to product accurate worldwide time series demand forecasts with a single model,
%been used by Amazon's Supply Chain Optimization Technologies (SCOT) team as a single model for forecasting time series worldwide.
which has enabled substantial model consolidation in global retail operations, improving forecast accuracy, while significantly reducing operational costs and associated technical debt.

%%However, as shown in section \ref{ss:architecture}, 
In spite of these successes, 
these and other neural network architectures continue to exhibit sub-optimal performance when training and evaluating on series with highly heterogeneous magnitudes \citep{wen2017mqrcnn, eisenach2020mqtransformer, wolff2024spade, salinas2019deepar, van2016wavenet, sutskever2014sequence, bai2018empirical}.
In demand forecasting for supply chain optimization, where average series magnitudes are often referred to as ``velocity'', this performance issue manifests as systematic forecasting bias and consequent accuracy degradation based on the velocity of the product; however, it is a ubiquitous phenomenon in many time series forecasting applications due to a combination of model architecture, training objective function, and training data distribution.
Since common objective functions and performance metrics implicitly favor high-magnitude series, biases against low-magnitude and sparse series not only exist, but can also be difficult to detect, e.g., as they are non-obvious through aggregate evaluation metrics.
As inventory management transitions to more granular level forecasting, the corresponding demand forecasting dataset becomes sparser. As the proportion of low-magnitude and sparse data increases, the cumulative bias can significantly impact overall forecasting accuracy.
See, e.g., Table \ref{tab:velocity_cat}: our ``moderate-velocity'' dataset (\texttt{DS3}, $\approx$3MM series) has as little as 9.62\% sparse series, 63.7\% of our ``low-velocity'' dataset (\texttt{DS1}, $\approx$700MM series) is sparse, and 90\% of ``extremely low-velocity'' dataset (\texttt{DS2}, $\approx$100MM series) is sparse. 
%

% https://code.amazon.com/packages/BatchEvaluationSpark/blobs/mainline/--/src/com/amazon/batch/evaluation/spark/precompute/dataset/VelocityDataset.scala
%    *ZERO: ${} = 0
%    * SUPER_SLOW: ${} BETWEEN (0, 2]
%    * SLOW: ${} BETWEEN (2, 52]
%    * MEDIUM: ${} BETWEEN (52, 365]
%    * FAST: ${} BETWEEN (365, 10000]
%     * SUPER_FAST: ${} > 10000
\begin{table}[t] %[h]
    \centering
    \begin{tabular}{llrrr}
        \toprule
        \textbf{Category} & \textbf{52-week agg.} & \textbf{\texttt{DS1}} & \textbf{\texttt{DS2}} & \textbf{\texttt{DS3}} \\
        \midrule
         Super Fast &  $(10000,\infty)$ & 0.05\% & 0.0007\% & 0.18\% \\
         Fast & $(365, 10000]$  & 1.6\% & 0.07\% & 11.77\% \\
         Medium & $(52,365]$  & 4.7\% & 0.4\% & 29.32\% \\
         Slow & $(2,52]$  & 17.3\% & 3.7\% & 43.40\% \\
         Super Slow & $(0,2]$  & 12.7\% & 5.7\% & 5.71\% \\
         Zero & $\{0\}$  & 63.7\% & 90\% & 9.62\% \\
         \midrule
         Approx. $N$ & (millions) & 700 & 100 & 3 \\
         \bottomrule
    \end{tabular}
    \caption{Magnitude categorization of target time series calculated by 52 trailing weeks/365 trailing days, and distribution across datasets. We refer to zero demand in the prior 52 weeks/365 days as ``extremely sparse~series.''}
    \label{tab:velocity_cat}
\end{table}

\begin{figure*}[t]
    \centering
    \includegraphics[width=.9\linewidth]{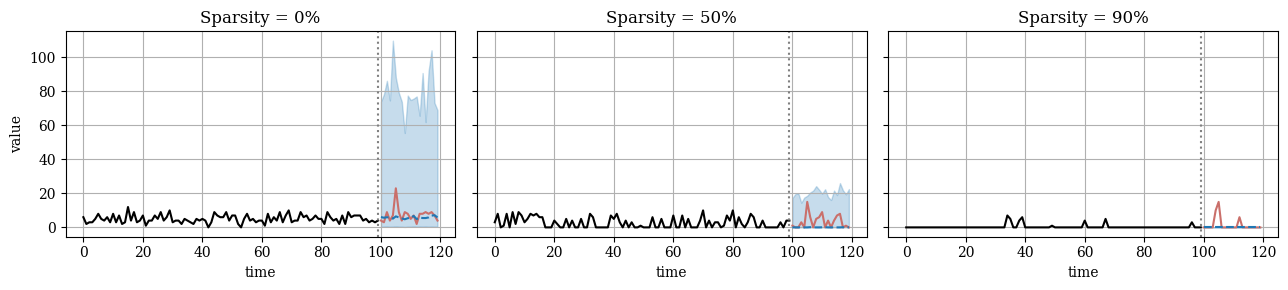}
    \caption{Convolutional forecast output by sparsity level; distributional collapse occurs at high levels of sparsity.}
    \label{fig:conv-sparse}
\end{figure*}

There are several related issues that arise with sparse and low-magnitude time series.
First, convolutional encoder-based models have limitations with extremely sparse time series, as they tend to collapse predictive distributions.
Figure \ref{fig:conv-sparse} illustrates such sparsity-induced under-dispersion, depicting the result of simulating forecasts from a single-layer causal CNN over histories whose zero fraction \(s\) is varied (the full sampling procedure is detailed in Appendix \ref{appendix:conv-collapse}).
As \(s\) rises from 0.0 to 0.9, the empirically estimated 80\% prediction band produced by the encoder narrows and collapses toward zero, illustrating the model’s shrinking uncertainty under extremely sparse inputs.
Second, convolutional encoders tend to integrate time series magnitudes inflexibly, resulting in biases dependent on the training data distribution;
for example, in \texttt{DS1}, high magnitude series exhibit significant under-bias when sampling randomly, and significant over-bias of the lowest magnitude series when sampling proportional to magnitude (see Appendix \ref{appendix:encoder-bias}).
Third, popular time series normalization techniques such as \texttt{RevIN} \citep{kim2022revin} are not well-compatible with forecasting tasks that requires a predictive distribution rather than a point forecast. 
%\RK{Is this claim mentioned in their/any reference paper (or) we observed it from our experiment?. We need some evidence to support this claim.} \MW{I'm not sure this needs too much evidence- it alludes to the fact that you can't just add back first and second moments into quantile forecasts. However, I do think Matthew did ablation experiments on this}

%%%
% Related Sparsity Work
%%%
There is a large and continuing body of work on forecasting sparse series \citep[e.g.,][]{Croston1972, syntetos2005accuracy, TEUNTER2011606, nikolopoulos2011aggregate, kourentzes2013intermittent, gutierrez2008lumpy, turkmen2019intermittent, zhang2024intermittent}.
%
%Perhaps most relevantly, in 2019, Ma and Forouzan \citep{ma2019amlc} introduced \texttt{Fjord}, a two-stage model which first predicted the probability of a zero forecast, and then used linear quantile regression to predict the parameters of a truncated shifted gamma (TSG) distribution to model aggregate future demand, leveraging parametric dis-aggregation properties of the gamma distribution to dis-aggregate to weekly demand. 
%
However, all of these methods share the core weakness that their objective function and internal scales operate in raw units, and do not adjust their training methodology nor their architecture to ensure accurate forecasts across all series magnitudes and sparsity levels.
These concerns are becoming particularly acute for development of a foundational model for arbitrary time series forecasting problems;
while the \texttt{SPADE} model has proven effective as a unified model for forecasting demand worldwide \citep{wolff2024spade}, expansion of the model to these additional use cases has exposed these emergent failure modes, and it is becoming critical that forecasting accuracy is independent of time series magnitudes.
These challenges highlight a need for new methods that effectively address complexities inherent in forecasting diverse and sparse time series data.

\paragraph{Main contributions.}
In this paper, we propose \texttt{SPADE-S}, a sparsity-robust model architecture that provides an effective architectural solution to forecasting heterogeneous series-magnitudes and sparsity levels.
Building upon the previously-developed \texttt{SPADE} forecasting model \citep{wolff2024spade}, we robustify the model to a wide range of series magnitudes and explicitly account for sparse time series, resulting in improved performance on these subsets without sacrificing overall accuracy, setting a new standard for reliable forecasting in retail and similar domains.
Our main contributions are the following.
\begin{enumerate}[(i)]
    \item 
    \textbf{Problem Characterization.}
    We identify several factors that lead existing models to systematically under-perform on low-magnitude and sparse time series, including loss functions with implicit biases against higher-magnitude series, sampling methods in the training, and normalization limitations of time series encoding methods.
    \item 
    \textbf{A Novel Time Series Encoder.} 
    We develop a novel multi-head dilated causal convolutional encoder module that provides critical flexibility to scale the architecture across highly diverse set of magnitudes in the time series data.
    \item 
    \textbf{Sparse Quantile Network.} 
    We develop a novel sparse model arm which uses a parametric distribution to more accurately represent the behavior of sparse series without distributional collapse.
    \item 
    \textbf{Generalized Accuracy Improvements at Scale.} 
    We demonstrate that our \texttt{SPADE-S} architecture is effective not only at scale, but robust to diverse use-cases---showing forecast improvements of up to 10.05\%, 14.80\%, and 6.10\% depending on forecasted quantile and time series magnitude, for \texttt{DS1}, \texttt{DS2}, and \texttt{DS3}, respectively; this results in respective overall P50 accuracy improvements of 0.92\%, 0.77\%, and 1.95\%, and overall P90 accuracy improvements of 2.21\%, 6.58\%, and 4.28\%, for these three use cases.
\end{enumerate}

% \section{Background}
% \input{kdd/sections/background}

\section{Methods}
In this section, we first describe the general forecasting task and it's consequences for forecasting across diverse time series; and
we then describe our architecture and it's novel contributions.
%

%%%
% National Forecasting Task
%%%

\subsection{Forecasting Task}
We consider a general product demand forecasting task \citep{wen2017mqrcnn,eisenach2020mqtransformer,wolff2024spade}, with forecast creation dates $t\in[T] \equiv \{1,\ldots,T\}$, products of interest $i \in\mathcal{I}$, and forecast horizons $h = (\text{lead-time}, \text{span})\in \mathcal{H}$ (a combination of approximately 240 valid lead-time/span pairs over the next 52 weeks). We'll denote the size of $\mathcal{H}$ as $|\mathcal{H}|$, and we'll denote all span-1 horizons as $\mathcal{H}_1$.
Our input covariates consist of past information $\mathbf{x}^{(p)}_{[t]} \in \mathbb{R}^{T\times d_p}$, known future information $\mathbf{x}^{(f)}_{[t],\mathcal{H}_1} \in \mathbb{R}^{T\times d_f \times |\mathcal{H}_1|}$, and static information $\mathbf{x}^{(s)} \in \mathbb{R}^{d_s}$, and we represent the target variable at time $t$ as $\mathbf{y}_{t,\mathcal{H}} \in \mathbb{R}^{T\times |\mathcal{H}|}$. 
The forecasting task estimates the following conditional distribution:
\begin{equation}
\mathbb{P}\left(\mathbf{y}_{t,\mathcal{H}}\,\mid\,
\btheta,\;
\mathbf{x}^{(p)}_{[t]},\; 
\mathbf{x}^{(f)}_{[t],\mathcal{H}_1},\;
\mathbf{x}^{(s)}\,\right).
\label{eq:normal_forecast}
\end{equation}

% Optimization Objective
%\paragraph{Optimization and Evaluation.}
%
We use weighted quantile loss (WQL) as our evaluation objective, where the WQL is defined as
\begin{align}
\label{equation:WQL}
\mathrm{WQL}(\mathbf{y}, \hat{\mathbf{y}}^{(q)}; q, \mathcal{I}, \mathcal{H}) &= \frac{\sum_{i,t,h}\QL\left(y_{i,t,h},\; \hat{y}^{(q)}_{i,t,h}(\btheta) ; q\right) } {\sum_{i,t,h}  y_{i,t,h} } ,
\end{align}
and $\QL\left(y,\; \hat{y};\ q\right)=q(y-\hat{y})_{+}+(1-q)(\hat{y}-y)_{+}$ is the quantile loss function, $\hat{y}^{(q)}$ denotes the estimated quantile, and $\btheta$ denotes a model in the class of models $\Theta$ defined by the model architecture.
We optimize $\btheta$ by minimizing the numerator of equation~\eqref{equation:WQL} summed across the quantiles of interest. See Appendix~\ref{appendix:training_methods} for details.

%%%
% Favoring high velocity
%%%
\paragraph{Magnitude-Bias of Common Loss Functions.}
Like most objectives, the Quantile Loss function introduces implicit inherent biases that warrant careful consideration.
To illustrate this, let \((y_{i,t,h},\hat y_{i,t,h})\) be the target and forecast for series \(i\), time \(t\), horizon \(h\).  
Assume every series is forecast with the same \emph{relative error}
\(r_{i,t,h}=r\) (e.g.\ a $10\%$ miss everywhere).
Then the absolute error equals
\(e_{i,t,h}=r\,y_{i,t,h}\).
Let the pointwise loss satisfy
\[
\ell(y,\hat y) \;=\; g(y)\,f(r), 
\qquad
r\equiv\frac{\hat y-y}{y},
\]
i.e.\ it factors into a scale term $g(y) > 0$ and a function $f(r) > 0$ of the relative
error $r$.  If $g$ is strictly increasing, then the global sample loss
\[
\mathcal L \;=\;\frac{1}{N}\sum_{i,t,h}\ell(y_{i,t,h},\hat y_{i,t,h})
\]
weights each series~$i$ in proportion to its total magnitude
\(
w_i \propto \sum_{t,h} g(y_{i,t,h}),
\)
so higher‑magnitude series contribute disproportionately (see Appendix \ref{proof:magnitude-bias} for the proof).
In particular, the quantile loss can be written as $|q - \mathbf{1}_{y<\hat{y}}||e| = |q-\mathbf{1}_{y<\hat{y}}||r|y = g(y)f(r)$ for $g(y) = y$.
Consequently, under equal relative error, time series with larger magnitudes exert strictly larger influence on the aggregate quantile loss.

Although prioritizing high-volume items can be reasonable in, e.g. supply-chain optimization, the implicit magnitude-based weighting becomes problematic when training unified, multi-purpose forecasters, if those forecasters do not have sufficient capacity to model heterogeneity in series-level magnitudes.
In product demand forecasting, for example, since the loss scales with absolute demand, (i) larger marketplaces, (ii) already successful products, and (iii) use-cases with inherently bigger signals all exert disproportionate influence, creating a self-reinforcing bias that can erode accuracy for emerging markets, new items, and low-scale applications—precisely the scenarios a truly universal model must handle equitably.
%\michael{This par probably goes in the intro. I can do it on the next iteration, once I get a better sense of what is going on, or you can do it and we iterate.}
% \RK{I think the most obvious question we can expect from reviewers might be: did we try any normalizing of loss functions if this is the fundamental problem we are addressing. Maybe, something we can also ablate for KDD paper but can skip for AMLC.} \MW{I agree we should do this as an ablation for KDD, but I don't think we have time to tackle this for AMLC} \RK{Yea.}
´

%%%
% SPADE-S
%%%
\subsection{SPADE-S Architecture}
\label{ss:architecture}

%\paragraph{\texttt{SPADE-S} Architecture.}
\begin{figure*}[t]
    \centering
    \includegraphics[width=.8\textwidth]{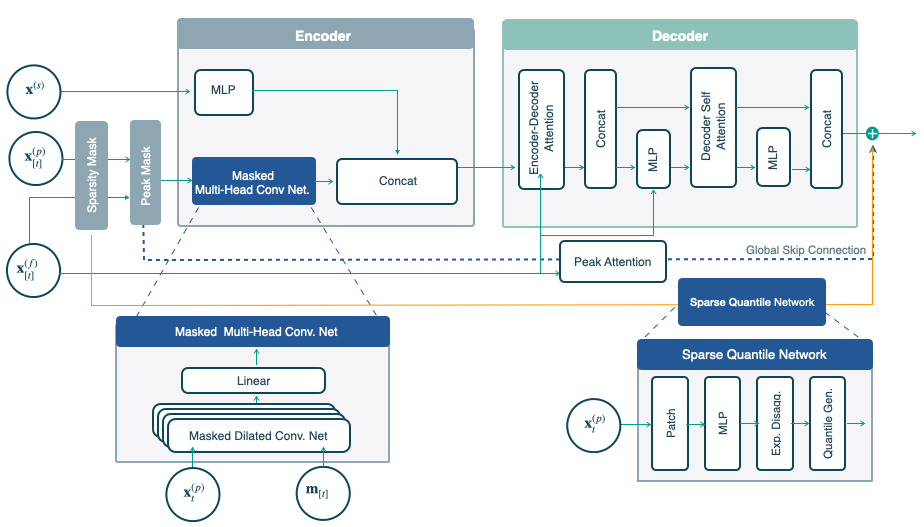}
    \caption{\texttt{SPADE-S} architecture, including modules to address diverse time series magnitudes, including masked multi-head dilated convolutional encoder, sparse series routing and a sparse quantile network.}
    \label{fig:arch}
\end{figure*}

Our model architecture, depicted in Figure \ref{fig:arch}, introduces novel contributions to address diverse time series magnitudes and sparsity levels, including a {masked multi-head dilated convolutional encoder}, {sparse series routing} and a {sparse quantile network}, which we describe below.

\paragraph{Multi-Head Convolutional Encoder.} 
Drawing inspiration from multi-head self-attention mechanisms \cite{vaswani2017attention} where different heads learn complementary input representations, we propose multi-head convolutional encoder that achieves similar representational diversity but with significantly lower computational overhead.

Parallel convolutional encoding has also been leveraged in other contexts: speeding up spectrogram inversion with parallel convolution, where each convolution learns a different interpolation pattern \citep{arik2018fast}; applying convolutional filters in parallel over each input series to condition on several covariates before combining them into a residual stack \citep{borovykh2017conditional}; and leveraging several parallel dilated convolutional heads with different rates in a multi-scale graph wavenet architecture for wind speed forecasting \citep{rathore2021multiscale}.

Unlike prior parallel convolution variants that handcraft each head’s receptive field or filter type, we show that by simply instantiating multiple identical dilated convolution stacks in parallel, we achieve the robustness and uncertainty benefits of an ensemble without manual head engineering or expensive training of separate models.

Specifically, we calculate
\begin{equation}
    \begin{aligned}
        \mathbf{e}_{[t],g}^{(p)} &= \text{Convolution}_g\left(\mathbf{\widetilde{x}_{[t]}^{(p)}}\right)\\
        %\text{Gate}_{[t]} &= \text{SoftMax}(\text{MLP}(\widehat{\mu}_{[t]},\widehat{\sigma}_{[t]}))\\
        \mathbf{e}_{[t]}^{(p)} &= \text{Linear}\left(\mathbf{e}_{[t],1}^{(p)},\ldots,\mathbf{e}_{[t],G}^{(p)}\right),
    \end{aligned}
\end{equation}
where $\widetilde{\mathbf{x}}_{[t]}^{(p)}$ are historical series inputs that have been peak-filtered (as in \texttt{SPADE} \citep{wolff2024spade}), and each individual head is a dilated causal convolution encoder seen in prior work \cite{wolff2024spade, eisenach2020mqtransformer, wen2017mqrcnn}.
Our method combines the variance‐reduction power of ensembles \citep{breiman1996bagging} with the efficiency of a single, shared‐structure encoder.

\paragraph{Sparse Series Routing.} 
To effectively route extremely sparse time series to a separate model arm, we leverage known information about the time series in a manner similar to \texttt{SPADE} \citep{wolff2024spade}.
We first separate the input batch into ``sparse" and ``non-sparse" series with the \texttt{SparsityMask} module.
We define ``sparse'' as having zero aggregate demand in the trailing 52 weeks, and not being classified as a new product, which we identify by date of first recorded product listing.
The ``non-sparse'' series are routed to the main encoder, which contains our masked multi-head convolutional encoder; and 
the ``sparse'' series are routed to our \texttt{SparseQuantileNetwork}.

\paragraph{Sparse Quantile Network.} 
The \texttt{SparseQuantileNetwork} first uses a patched MLP to estimate the parameters of a simple parametric distribution, dis-aggregates those parameters across the horizons in $\mathcal{H}$, and uses the horizon-specific parameters to produce a quantile forecast via an inverse cumulative distribution function (ICDF).
Note that this general frame works for any simple parameteric distribution; one could estimate parameters from an, e.g., Truncated Shifted Gamma (TSG) distribution and dis-aggregate using it's additive properties (see Appendix \ref{appendix:tsg}). However, the ICDF of a Gamma has no closed form solution, requiring sample path generation which significantly increases training and inference costs for tail quantiles.
One could also use the additivity of i.i.d. gaussian parameters to generate quantiles for a truncated normal (see Appendix \ref{appendix:tsg})---results we illustrate in our ablation study (Appendix \ref{appendix:ablation}).

However, to retain both computational simplicity as well as to capture the heavy-tail characteristics of empirical product demand distributions, we assume the quantiles across horizons $h$ to come from a simpler exponential distribution, whose simple ICDF is 
$\hat{y}_{[t],\mathcal{H}}^{(q)} = -  h(\text{span},\vartheta_{[t]})\ln(1-q),$
where $h(\text{span}, \vartheta_{[t]}) = \frac{\text{span}}{\varsigma}\vartheta_{[t]}$ represents the exponential scale parameter, assuming that span-1 demand variables are perfectly correlated.
To summarize, the sparse model arm does the following:
\begin{equation}
\begin{aligned}
    \mathbf{s}_{[t]}^{(p)} &= \text{Patch}\left(\mathbf{x}_{[t]}^{(p)}\right)\\
    \vartheta_{[t]} &= \text{MLP}\left(\mathbf{s}_{[t]}^{(p)}\right)\\
    % \hat{y}_{[t],\mathcal{H}}^{(q)} &= -  \frac{\text{span}}{\varsigma}\vartheta_{[t]}\ln(1-q)
    \hat{y}_{[t], \mathcal{H}}^{(q)} &= F^{-1}\left(q;\ h(\text{span},\ \vartheta_{[t]}\right)),
\end{aligned}    
\end{equation}
where $h(\text{span}, \vartheta_{[t]})$ is a function of span, and $ $$F^{-1}(q)$ denotes the quantile function for the particular parametric distribution used.
Restriction of the forecast to a simple parametric distribution is critical in robust estimation of sparse quantile forecasts. In this case, we always forecast quantiles no more than P50 to be zero, as any continuous distribution would result in P50 over-bias, and learned probability masking results in difficult-to-learn architectures using discrete optimization methods. 
%{\color{red}Ruijun: quick comment -- i think the real reason this version is better than previous ones using TSG and TruncNormal is that this one disaggregates the scale parameter, which assumes perfect correlation among span1s. With this assumption, given P90 in the maximum span horizon, all P90 quantiles for lower-spans will be minimized, and being pulled towards 0. For the two alternative models we experimented in the appendix, both probabilistic models are under the i.i.d. span1 demand assumption, this will make lower-span p90s higher. That being said, I think we may get similar or better results if we disaggregate scale parameter for TSG (Exponential is a special case for Gamma).}
%\MW{Great point. Let's plan to do an ablation on the TSG based on this. I'd actually prefer to use the TSG, because the exponential will likely not be as good for the Span N forecasts.}

\section{Results}
\label{sec:results}
%%%
% 
%%%
We evaluate \texttt{SPADE-S} on three diverse large-scale product forecasting applications.
See Table \ref{tab:velocity_cat} for a summary.
Each of these datasets not only have different distribution of series magnitudes (see, e.g., Table \ref{tab:velocity_cat}), but also variable input and product demand characteristics, which we describe below.
In what follows, we'll review the experimental setup of each of these applications, including their particular data and forecasting task, and then we will present our main results.

\subsection{Setup of Our Empirical Evaluation}

\paragraph{Low velocity series forecasts (\texttt{D1}).}
Our first use case is world-wide listing level forecasts for online retail products.
In this use case, we aim to predict listing demand across the entire forecast horizon $\mathcal{H}$.
The training data is weekly grain information spanning 260 weeks from 2017 to 2022 and consisting of hundreds of millions of unique listings.
The backtest data consists of the subsequent 52 weeks in 2023 after the training period.
The time series in the backtest are a total population of nearly one billion unique listing series.

\vspace{-2mm}
\paragraph{Extremely low velocity forecasts (\texttt{D2}).}
%https://quip-amazon.com/VCC8AuXsQ3FO/KDD-Sparsity-Regional
Our second use case is forecasting weekly product demand for an online retailer per geographic area.
%\MW{Can we define these things, here or in the appendix?}
Unlike the prior use case, the relevant forecast is up to a 10 week horizon.
The training data is weekly grain information spanning 260 weeks from 2017 to 2022 of over 100 million series. 
Similar to \texttt{D1}, we include features to capture exogenous information such as holidays and promotions. 
For the backtest data, we use a uniform random sample of over 100 million series for the subsequent 52 weeks in 2023 after the training period.
As is shown in Table \ref{tab:velocity_cat}, more than $90\%$ of the series are sparse (categorized as ``Zero''), while fewer than $1\%$ of the series are considered ``Medium'' or faster.

\vspace{-2mm}
\paragraph{Moderate velocity forecasts \texttt{(D3)}.}
Our third use case is forecasting daily demand for products at the store level.
In this use-case, we aim to forecast across an entire forecasting horizon $\mathcal{H}$ = 91 days and various spans ranging consisting of a total of 285 lead-time/span combinations.
Since the product selection for stores is limited compared to online marketplaces, it is very unlikely to retain offers for the slowest products, and hence the \% of super-fast and fast categories are much higher up to 12\% vs <2\% for other use-cases.
The training data consist of daily grain information spanning 730 days, starting from 2022 to 2023 and totaling over 3 million unique series.

The dataset also include features to capture exogenous information such as holidays and promotions along with static information about the product like brand, category etc. 
The backtest data consist of subsequent 365 days after the end of training period for each product-store time series starting from 2024.

\subsection{Main Empirical Results}

Our main empirical results are presented in Table \ref{table:main}.
Results are displayed as the percent difference in P50 and P90 quantile loss relative to the baseline model.
The baseline models are \texttt{SPADE} \citep{wolff2024spade} for \texttt{D1} and \texttt{D2}, and \texttt{MQTransformer} \citep{eisenach2020mqtransformer} for \texttt{D3}\footnote{\texttt{D3} does not observe the same extreme holiday- and promotion-related spikes as other demand forecasting problems, so \texttt{PeakAttention} is unnecessary.}.
Results are decomposed by the time series magnitude categorizations shown in Table \ref{tab:velocity_cat}, along with the proportion of evaluation data falling into each of these categorizations. 

\begin{table}[htbp]
\centering
\begin{tabular}{rrccc}
    \toprule
    \bfseries Category & \bfseries Metric 
      & \bfseries \texttt{D1} & \bfseries \texttt{D2} & \bfseries \texttt{D3} \\
    \midrule
    \multirow{2}{*}{All}
      & P50 & -0.92\% & -0.77\% & -1.95\%   \\
      & P90 & -2.21\% & -6.58\% & -4.28\%  \\
    \addlinespace
    \multirow{2}{*}{Super Fast}
      & P50 & -0.77\% & -2.00\% & -1.94\%  \\
      & P90 & -1.20\% & -10.30\% & -3.28\%  \\
    \addlinespace
    \multirow{2}{*}{Fast}
      & P50 & -0.72\% & -3.90\% & -2.99\%  \\
      & P90 & -1.02\% & -14.80\% & -6.10\%  \\
    \addlinespace
    \multirow{2}{*}{Medium}
      & P50 & -1.01\% & -2.50\% & -1.03\%  \\
      & P90 & -1.38\% & -12.20\% & -2.52\%  \\
    \addlinespace
    \multirow{2}{*}{Slow}
      & P50 & -0.95\% & 1.60\% & -0.01\%  \\
      & P90 & -2.50\% & -3.60\% & -0.51\%  \\
    \addlinespace
    \multirow{2}{*}{Super Slow}
      & P50 & -0.94\% & 0.80\% & -0.01\%  \\
      & P90 & -5.54\% & -0.30\% & -1.03\%  \\
    \addlinespace
    \multirow{2}{*}{Zero}
      & P50 & -4.37\% & 0.40\% & -0.45\%  \\
      & P90 & -10.05\% & 0.20\% & 0.27\%  \\
    \bottomrule
  \end{tabular}
  \caption{Main results of \texttt{SPADE-S} by task and series magnitudes defined in Table \ref{tab:velocity_cat}, compared to benchmark models.}
  \label{table:main}
\end{table}

As shown in Table \ref{table:main}, we find general improvement across all magnitude categories and use-cases.
Moreover, our magnitude level results suggest that baseline models tend to favor the construction of accurate forecasts for the highest magnitude targets to the detriment of lower magnitude targets.
\texttt{SPADE-S} alleviates this issue.
For \texttt{D1}, \texttt{SPADE-S} notably shows P90 forecast improvements on ``slow'', ``super slow'', and ``zero'' products of 2.50\%, 5.54\%, and 10.05\%, respectively, and a P50 forecast improvement of 4.37\% for ``zero'' products---improvements primarily driven by significant reductions in over-bias of the forecast, as seen in Appendix Table \ref{table:ablation}.  

For \texttt{D2}, we also observe significant improvement on P90 forecast---10.30\% on ``Super Fast'', 14.80\% on ``Fast'', and 12.20\% on ``Medium'' series. Sparse series routing prevented zero-value products from biasing faster-moving products, reducing under-prediction and improving overall accuracy. Detailed results are in Appendix \ref{appendix:regional}.

For \texttt{D3}, we also observe large improvements in high magnitude time series categories---with P90 improvements of 3.28\%, 6.10\% and 2.52\% in ``super fast'', ``fast'', and ``medium'' categories, primarily driven by reduction in the under-bias of the forecast. Since the number of zero magnitude time series is relatively less (i.e, <10\%) compared to national/regional use-cases (i.e., >63\% to 90\%), routing them to sparse ARM is not necessary to achieve overall improvements in both high and low magnitude time series categories.

% \subsection{Routing mechanisms in Masked Convolution Mixture-of-Experts Encoder}

% In this section, we explore various routing mechanisms of the masked convolution mixture-of-experts (MoE) encoder and their impact on the model performance and training efficiency. We analyze the two key aspects of routing in MoE encoder: 1. The choice between learned routing and domain-knowledge based routing indicating if the selection of expert is learned based on optimizing the overall objective function (or) selected using heuristic rules based on domain knowledge. 2. The choice between hard and soft routing indicating whether only one expert (or) weighted ensemble of experts is selected for encoding an input time series.

% \textbf{Learned routing vs domain-knowledge based routing} In learned routing, the router learns to map the time series scale (mean, std) to one of the experts through end-to-end optimization via weighted quantile loss. This allows the router to learn complex patterns based on scale for expert selection but is less interpretable. On the other hand, the domain-knowledge based router maps the time series scale (mean) to one of the experts via the velocity groups described in Table \ref{tab:velocity_cat} and gives supervision for each of the expert to specialize in encoding specific velocity group.

% \textbf{Soft routing vs hard routing.} 

\section{Discussion}
\texttt{SPADE-S} is a significant advancement in addressing heterogeneous time series data characterized by varying magnitudes and sparsity patterns.
Empirical results on three separate massive internal datasets reveal several key insights worth examining.
Most notably, the substantial improvements in forecast accuracy across different use cases validate our initial hypothesis that existing models systematically under-perform on low-magnitude and sparse time series. 
In forecasting across nearly 1 billion series in \texttt{D1}, \texttt{SPADE-S} achieved up to 10\% improvement in accuracy, depending on the quantile forecast and magnitude of the series. 
Similarly, forecasting across over 100 million series and over 3 million series in \texttt{D2} and \texttt{D3} showed improvements of up to 15\% and 6\% respectively.
These results suggest that the model's benefits are transferable across varying dataset size and complexity.
Moreover, P90 accuracy gains of 2.21\% for \texttt{D1}, 6.58\% for \texttt{D2}, and 4.28\% for \texttt{D3} indicate that \texttt{SPADE-S} handles tail estimation more effectively than existing approaches, while the P50 gains of 0.92\% and 0.77\% and 1.95\% respectively suggest that the model maintains strong performance across median cases.

These findings have significant practical implications for industries relying on large-scale time series forecasting, particularly in retail and supply chain management.
The ability to forecast more accurately across varying magnitudes and sparsity patterns could lead to improved inventory management, reduced waste, and more efficient resource allocation.
Moreover, the success of our multi-head convolutional encoder opens new avenues for research in handling heterogeneous multivariate time series, and this architectural innovation could potentially be adapted for other applications beyond demand forecasting.
%
%Other potential research directions include a systematic study of depth–width trade-offs in deep multi-head convolutional architectures---e.g. comparing performance gains from stacking additional CNN blocks versus widening each layer’s channel capacity---and the derivation of empirical scaling laws that quantify how forecasting accuracy scales with overall model parameter count.

\section{Citations}
\bibliographystyle{plain}
\bibliography{main}

% \section{Acknowledgments}

% ...
\newpage
\section*{Appendices}

% %%
% %% If your work has an appendix, this is the place to put it.
\appendix
\section{Supplementary Details}
\subsection{Convolutional collapse simulation details.}
\label{appendix:conv-collapse}
We construct a synthetic experiment to illustrate how overall sparsity degrades the predictive dispersion produced by a convolutional encoder.
For each sparsity level \(s\in\{0,0.5,0.9\}\) we generate a 100-step history \(y_{1{:}100}\) from a Poisson\((5)\) process and randomly set a fraction \(s\) of its entries to zero.
The history is fed into a single-layer causal CNN with a 24-lag, two-channel, exponentially-weighted kernel whose coefficients are normalised to sum to one; a learned scalar bias is added to the convolutional output. 
Letting \(\mu\) denote the last convolved value and \(\sigma\) the standard deviation of the most recent 30 non-zero observations, we draw 500 Monte-Carlo sample paths for the 20-step forecast horizon according to \(\hat y_{t+h}^{(d)}=\mu+\sigma z_{h}^{(d)}\) with \(z_{h}^{(d)}\sim\mathcal N(0,1)\). 
The empirical 10th, 50th and 90th percentiles across the draws yield an 80\% prediction interval \([\hat y^{(10)},\hat y^{(90)}]\) whose width—and ultimately its collapse toward zero—is visualised as sparsity increases. 
A future trajectory with the same sparsity pattern is overlaid to highlight the encoder’s growing under-dispersion and its failure to capture demand resurgence.

\subsection{Bias Trade-off by Sampling Scheme}
\label{appendix:encoder-bias}

\begin{figure}[h]
    \centering
    \includegraphics[width=\linewidth]{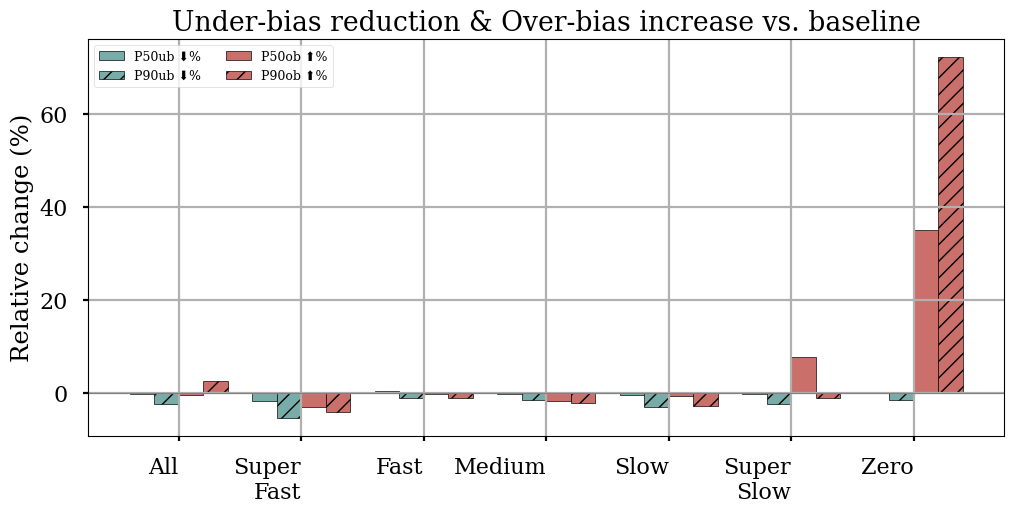}
    \caption{Relative change of under-bias and over-bias versus baseline forecast by sampling scheme for D1 dataset.}
    \label{fig:bias-by-sampling}
\end{figure}

To show the under-bias over-bias trade-off by sampling scheme, we use \texttt{D1} training and backtest data described in section \ref{sec:results}.
We train a baseline \texttt{SPADE} \citep{wolff2024spade} model, which serves as the same baseline as our main results, using series magnitude-based importance sampling with a cutoff quantile of 0.8---i.e., we sample more frequently in proportion to series magnitude for products with magnitudes above the P80 quantile, and use uniform sampling below P80, where the uniform weight is equivalent to a P80 magnitude.
Our experimental model trains the same \texttt{SPADE} architecture, but uses series magnitude based importance sampling for the entire magnitude distribution---i.e., all products are sampled according to their series magnitudes.
Figure \ref{fig:bias-by-sampling} shows that the experimental model, which samples according to velocity across the entire distribution, shows over-bias on P90 for ``super slow'' products, and extreme over-bias on at P90 for ``zero'' products, while improving on both under-bias and over-bias for the remaining products. However, the extreme over-bias on ``zero'' products, given that they make up for over 60\% of the total population, results in overall over-bias of the model.

% \subsection{Lead-Time Span Pairs}
% \label{appendix:ltsps}
% Figure \ref{fig:ltsps} depicts the lead-time span pairs used in this multi-horizon forecasting problem.

% \begin{figure}[h]
%     \centering
%     \includegraphics[width=\linewidth]{STALE_kdd/img/ltsps.png}
%     \caption{Lead-Time/Span pairs included in this multi-horizon forecasting problem.}
%     \label{fig:ltsps}
% \end{figure}

\subsection{Training Methods}
\label{appendix:training_methods}

Let $\btheta$ be a model that resides in the class of models $\Theta$ defined by the model architecture. Let $\mathcal{A}$ the dataset's products, and $\mathcal{H}$ the horizon defined by lead times and spans. We train a quantile regression model by minimizing the following multi-quantile loss:
\begin{align}
\label{equation:learning_objective}
    \min_{\btheta}
    \sum_{q} \sum_{i} \sum_{t} \sum_{h}  \QL\left(y_{i,t,h},\; {\hat{y}}^{(q)}_{i,t,h}(\btheta) ;\ q\right),
\end{align}
%\michael{bit more detail to be self contained, e.g., QL is such-and-such, and that we have a LT-span grid that defined some of the sums.}
for products $i\in \mathcal{I}$, time $t$ and horizon $h\in\mathcal{H}$, and $\underline{\hat{y}}^{(q)}$ denotes the estimated quantile\footnote{During training, demand and forecasts are normalized by the length of the horizon $h$.}. We optimize \texttt{SPADE-S} using stochastic gradient descent with \emph{Adaptive Moments} (\texttt{ADAM}; \cite{kingma2014adam_method}). %, and the subscript $+$ means the positive part.

\subsection{Magnitude-Bias of Common Loss Functions}
\label{proof:magnitude-bias}
\paragraph{Proposition.}
Let \((y_{i,t,h},\hat y_{i,t,h})\) be the target and forecast for series \(i\), time \(t\), horizon \(h\).  
Assume every series is forecast with the \emph{same relative error}
\(r_{i,t,h}=r\) (e.g.\ a $10\%$ miss everywhere).
Then the absolute error equals
\(e_{i,t,h}=r\,y_{i,t,h}\).
Let the pointwise loss satisfy
\[
\ell(y,\hat y) \;=\; g(y)\,f(r), 
\qquad
r:=\frac{\hat y-y}{y},
\]
i.e.\ it factors into a scale term $g(y) > 0$ and a function $f(r) > 0$ of the relative
error $r$.  If $g$ is strictly increasing, then the global sample loss
\[
\mathcal L \;=\;\frac{1}{N}\sum_{i,t,h}\ell(y_{i,t,h},\hat y_{i,t,h})
\]
weights each series~$i$ in proportion to its total magnitude
\(
w_i \propto \sum_{t,h} g(y_{i,t,h}),
\)
so higher‑magnitude series contribute disproportionately.

\begin{proof}
Insert $\ell=g(y)f(r)$ into $\mathcal L$ and regroup by series:
\[
\mathcal L
  =\frac{1}{N}\sum_i\sum_{t,h}g(y_{i,t,h})f(r_{i,t,h})
  =\sum_i\Bigl(\tfrac{\sum_{t,h}g(y_{i,t,h})}{N}\Bigr)
        \Bigl(\tfrac{\sum_{t,h}g(y_{i,t,h})f(r_{i,t,h})}
                   {\sum_{t,h}g(y_{i,t,h})}\Bigr).
\]
The first factor is the series weight \(w_i\), which grows with~$y$ because
$g$ is increasing, proving the bias.
\end{proof}

We examine three common losses in this regime.
%--------------------------------------------------------------------
\paragraph{1.  MSE (Mean–Squared Error).}

\[
\ell_{\mathrm{MSE}}
      \;=\;(\hat y-y)^2
      \;=\;e^{2}
      \;=\;r^{2}\,y^{2}
      \quad\Longrightarrow\quad
      g(y)=y^{2}.
\]

Thus, under equal relative error its effective weight grows quadratically with magnitude, so large‑scale series dominate the summed loss.

%--------------------------------------------------------------------
\paragraph{2.  CRPS (Continuous Ranked Probability Score).}

For many location‑scale forecast families (e.g.\ Normal, Laplace) one can
show
\[
\mathrm{CRPS}(F,y)
      \;=\;\sigma\,\varphi(r)
      \quad\text{with}\quad
      r=\frac{\hat\mu-y}{\sigma},
\]
where $\sigma$ is the predictive scale and $\varphi$ depends only on the standardised error.  If forecasts keep afixed relative spread,
$\sigma=\kappa\,y$, then
\[
\mathrm{CRPS} \propto y,
\quad\text{i.e.}\quad g(y)=y,
\]
again privileging high‑magnitude series.

%--------------------------------------------------------------------
\paragraph{3.  Quantile (Pinball) Loss.}

For a $\tau$‑quantile forecast $\hat q$, the pinball loss is
\(
\ell_\tau = ( \tau-\mathbf 1_{y<\hat q})(y-\hat q)
          = |\tau-1_{y<\hat q}|\,|e|.
\)
Equal relative error implies $|e|=|r|y$, giving \(g(y)=y\).

%--------------------------------------------------------------------

\subsection{Sparse Quantile Network Parametric Distributions.}
\label{appendix:tsg}
\paragraph{Gamma}
For i.i.d. Gamma variables $X_i \sim \text{Gamma}(k_i, \vartheta)$, 
$$
\sum_i X_i \sim \mathrm{Gamma}\!\Bigl(\sum_i k_i,\;\vartheta\Bigr),
$$
We can first estimate the scale $\vartheta$ for the maximal forecasted span $\varsigma$ in the horizon set $\mathcal{H}$, and then disaggregate the distribution under the assumption the shape parameter is proportional to span, using the associated distribution to generate a quantile forecast. However, the inverse quantile function does not have a closed form, which necessitates generation of many sample paths to produce a backpropagatable quantile estimate.

\paragraph{Truncated Normal.}
If one instead assumes \(X_1,\dots,X_n\stackrel{\mathrm{iid}}{\sim}\mathcal N(\mu,\sigma^2)\). Then
\[
S=\sum_{i=1}^n X_i\sim\mathcal N\bigl(n\mu,\;n\sigma^2\bigr),\quad
\bar X=\frac{1}{n}S\sim\mathcal N\Bigl(\mu,\;\frac{\sigma^2}{n}\Bigr).
\]
Thus aggregation of \(n\) iid normals gives
\[
\mu_S = n\mu,\qquad \sigma^2_S = n\,\sigma^2,
\]
and dis‐aggregation (recovering the original parameters) is
\[
\mu = \frac{\mu_S}{n},\qquad \sigma^2 = \frac{\sigma_S^2}{n}.
\]
This can be leveraged to decompose estimated parameters by span. Then the inverse quantile function is
\[
\text{For }X\sim\mathcal N(\mu,\sigma^2),\;F_X(x)=\Phi\bigl(\tfrac{x-\mu}{\sigma}\bigr)
\quad\Longrightarrow\quad
F_X^{-1}(p)=\mu+\sigma\,\Phi^{-1}(p),
\]
\[
\Phi^{-1}(p)=\sqrt{2}\,\mathrm{erf}^{-1}(2p-1),
\]
and the result can be turned into a truncated normal with a ReLU on the quantile forecast produced.

\section{Ablation Studies}\label{appendix:ablation}
We run a number of architectural ablation studies to compare different methodologies.
All ablations follow the same experimental pipeline as our main results.
In the ablations below, \texttt{Spade V0} refers to the original \texttt{SPADE} architecture \citep{wolff2024spade}.
``Adjusted cutoff quantile'' is a parameter of our training sampling scheme, which uses magnitude-based importance sampling; the cutoff quantile is the quantile in which every observation below the magnitude $q$ receives uniform weight, with weight equivalent to the quantile $q$.
``Rule-based'' override layer refers to a \texttt{Sparse Quantile Network} that forces all sparse series to forecast 0 for P50 and 0 for P90.
MoE attempts to learn the encoder head mixture through the average mean and variance of the historic target values. Poseterior analysis reveals that the learned weight is approximately uniform.
The learned P90 layers experiment with multiple different distributions in the \texttt{SparseQuantileNetwork.}
\begin{enumerate}
    \item[\textbf{V9.}] Spade V0 
    \item[\textbf{V10.}] Spade V0 + Adjusted cutoff quantile (0.8 to 0.1) 
    \item[\textbf{V11.}] Spade V0 + Rule-based override layer
    \item[\textbf{V12.}] Spade V0 + Adjusted cutoff quantile (0.8 to 0.1) + Rule-based override layer
    \item[\textbf{V13.}] Spade V0 + MoE with soft routing (6 experts)
    \item[\textbf{V15.}] Spade V0 + learned P90 layer (truncated normal distribution on raw input)
    \item[\textbf{V16.}] Spade V0 + MoE (6 experts) + learned P90 layer (truncated normal distribution on raw input)
    \item[\textbf{V17.}]  Spade V0 + learned P90 layer (exponential distribution on raw input)
    \item[\textbf{V18.}] Spade V0 + MoE (6 experts) + learned P90 layer (exponential distribution on raw input)
    \item[\textbf{V19.}] Main Model (Figure \ref{fig:arch})
\end{enumerate}

\subsection{\texttt{D1} Ablations}
Table \ref{table:ablation} shows detailed experimental results on the D1 dataset.
We find that V19, the main model, is the consistently most successful model across all velocities and overall.
It's worth noting that V13, which has an MoE encoder without a sparse arm, has nearly competitive performance outside of the ``Zero'' category, but the sparse arm has an important effect on this category that influences the overall quantile loss, given it makes up for over 60\% of the dataset.

\begin{table*}[t]
\resizebox{\textwidth}{!}{%
\begin{tabular}{rlcccccccccc}
    \toprule
    &            & \multicolumn{10}{c}{\bfseries Version} \\
    \addlinespace
    \bfseries Category & \bfseries Metric 
      & \bfseries V9 & \bfseries V10 & \bfseries V11 
      & \bfseries V12 & \bfseries V13 & \bfseries V15 
      & \bfseries V16 & \bfseries V17 & \bfseries V18 & \bfseries V19 \\
    \midrule
    \multirow{6}{*}{All}
      & P50ql & & 0.04\% & -0.42\% & -0.46\% & -0.58\% & -0.29\% & -0.50\% & -0.38\% & {-0.79\%} & \textbf{-0.92\%} \\
      & P90ql & & 2.38\% & -1.33\% & -1.38\% & -1.33\% & -0.22\% & -1.44\% & -1.27\% & {-1.94\%} & \textbf{-2.21\%} \\
    \addlinespace
      & P50ob & & -0.54\% & -4.90\% & -0.12\% & -3.63\% & -1.15\% & 0.12\% & -2.24\% & -6.65\% & -2.18\% \\
      & P50ub & & 0.32\% & 1.88\% & -0.67\% & 0.99\% & 0.13\% & -0.89\% & 0.54\% & 2.29\% & -0.32\% \\
      & P90ob & & 2.41\% & -5.57\% & -4.26\% & -4.05\% & -2.16\% & -2.29\% & -5.03\% & -8.04\% & -4.81\% \\
      & P90ub & & 2.41\% & 3.72\% & 2.08\% & 1.97\% & 2.19\% & -0.33\% & 3.29\% & 5.37\% & 0.99\% \\
    \cmidrule{2-12}
    \multirow{6}{*}{Super Fast}
      & P50ql & & -0.06\% & 0.13\% & -0.26\% & {-0.51\%} & 0.06\% & -0.26\% & 0.13\% & -0.26\% & \textbf{-0.77\%} \\
      & P90ql & & 0.22\% & 0.00\% & -0.33\% & {-0.98\%} & 0.00\% & -0.44\% & 0.00\% & -0.65\% & \textbf{-1.20\%} \\
    \addlinespace
      & P50ob & & -3.14\% & -4.66\% & 0.17\% & -2.46\% & -1.86\% & 0.51\% & -2.63\% & -4.75\% & -0.51\% \\
      & P50ub & & 1.79\% & 3.07\% & -0.51\% & 0.67\% & 1.28\% & -0.72\% & 1.79\% & 2.46\% & -0.87\% \\
      & P90ob & & -4.12\% & -2.98\% & -1.99\% & -1.40\% & -1.83\% & -1.38\% & -2.86\% & -4.20\% & -1.93\% \\
      & P90ub & & 5.49\% & 3.74\% & 1.76\% & -0.44\% & 2.42\% & 0.66\% & 3.74\% & 3.74\% & -0.22\% \\
    \cmidrule{2-12}
    \multirow{6}{*}{Fast}
      & P50ql & & -0.38\% & -0.30\% & -0.34\% & -0.55\% & -0.26\% & -0.38\% & -0.34\% & \textbf{-0.72\%} & \textbf{-0.72\%} \\
      & P90ql & & -0.32\% & -0.32\% & -0.51\% & -0.83\% & -0.32\% & -0.76\% & -0.19\% & {-0.89\%} & \textbf{-1.02\%} \\
    \addlinespace
      & P50ob & & -0.28\% & -3.73\% & 1.07\% & -3.27\% & -0.56\% & 0.56\% & -1.30\% & -6.78\% & -0.96\% \\
      & P50ub & & -0.44\% & 1.74\% & -1.23\% & 1.06\% & -0.07\% & -0.96\% & 0.24\% & 2.90\% & -0.55\% \\
      & P90ob & & -1.27\% & -2.13\% & -1.36\% & -2.19\% & -1.57\% & 0.16\% & -1.95\% & -5.74\% & -0.79\% \\
      & P90ub & & 1.07\% & 2.13\% & 0.67\% & 1.07\% & 1.46\% & -1.86\% & 2.26\% & 5.73\% & -1.20\% \\
    \cmidrule{2-12}
    \multirow{6}{*}{Medium}
      & P50ql & & -0.44\% & -0.69\% & -0.57\% & -0.66\% & -0.41\% & -0.69\% & -0.47\% & \textbf{-1.23\%} & -1.01\% \\
      & P90ql & & -0.49\% & -0.45\% & -0.89\% & -1.02\% & -0.89\% & -1.30\% & -0.45\% & \textbf{-1.58\%} & -1.38\% \\
    \addlinespace
      & P50ob & & -1.79\% & -3.86\% & 0.37\% & -4.28\% & 1.38\% & 1.79\% & -0.28\% & -6.67\% & -2.16\% \\
      & P50ub & & 0.26\% & 1.01\% & -1.03\% & 1.25\% & -1.30\% & -1.97\% & -0.55\% & 1.63\% & -0.38\% \\
      & P90ob & & -2.14\% & -2.60\% & -1.67\% & -3.33\% & -1.88\% & 0.65\% & -0.95\% & -5.48\% & -1.69\% \\
      & P90ub & & 1.59\% & 2.27\% & 0.08\% & 1.93\% & 0.34\% & -3.86\% & 0.17\% & 3.36\% & -1.01\% \\
    \cmidrule{2-12}
    \multirow{6}{*}{Slow}
      & P50ql & & 0.15\% & -0.57\% & {-0.62\%} & -0.17\% & -0.37\% & -0.40\% & -0.55\% & \textbf{-0.62\%} &\textbf{ -0.95\%} \\
      & P90ql & & -0.08\% & -1.18\% & -1.84\% & -1.61\% & -2.02\% & \textbf{-2.94\%} & -1.84\% & -1.97\% & -2.50\% \\
    \addlinespace
      & P50ob & & -0.67\% & -6.97\% & -3.53\% & -2.34\% & -2.43\% & -0.33\% & -3.77\% & -6.40\% & -4.44\% \\
      & P50ub & & 0.44\% & 1.67\% & 0.41\% & 0.59\% & 0.34\% & -0.42\% & 0.59\% & 1.42\% & 0.29\% \\
      & P90ob & & -2.90\% & -6.74\% & -6.37\% & -6.36\% & -7.01\% & -6.91\% & -7.28\% & -8.77\% & -7.17\% \\
      & P90ub & & 3.04\% & 4.93\% & 3.19\% & 3.62\% & 3.48\% & 1.40\% & 4.16\% & 5.51\% & 2.66\% \\
    \cmidrule{2-12}
    \multirow{6}{*}{Super Slow}
      & P50ql & & 1.53\% & -0.63\% & -0.29\% & 0.08\% & -0.67\% & 0.36\% & {-0.77\%} & -0.25\% & \textbf{-0.94\%} \\
      & P90ql & & 0.75\% & -4.07\% & -5.04\% & -3.51\% & -4.91\% & -4.98\% & {-5.20\%} & -4.58\% & \textbf{-5.54\%} \\
    \addlinespace
      & P50ob & & 7.56\% & -13.87\% & -4.37\% & -6.79\% & -8.44\% & -0.94\% & -10.57\% & -11.87\% & -12.51\% \\
      & P50ub & & 0.22\% & 2.23\% & 0.56\% & 1.54\% & 1.01\% & 0.62\% & 1.31\% & 2.23\% & 1.54\% \\
      & P90ob & & -1.13\% & -15.49\% & -15.21\% & -13.02\% & -15.29\% & -13.69\% & -16.19\% & -17.16\% & -17.78\% \\
      & P90ub & & 2.49\% & 6.44\% & 4.31\% & 5.23\% & 4.62\% & 3.04\% & 4.89\% & 6.99\% & 5.72\% \\
    \cmidrule{2-12}
    \multirow{6}{*}{Zero}
      & P50ql & & 6.16\% & -3.50\% & -1.64\% & -2.68\% & -2.39\% & {-3.68\%} & -3.02\% & -3.19\% & \textbf{-4.37\%} \\
      & P90ql & & 32.57\% & \textbf{-10.36\%} & -6.76\% & -3.71\% & 7.17\% & -2.73\% & -8.81\% & -9.61\% & \textbf{-10.05\%} \\
    \addlinespace
      & P50ob & & 35.06\% & -28.00\% & -14.44\% & -22.11\% & -19.85\% & -26.07\% & -24.59\% & -25.96\% & -32.91\% \\
      & P50ub & & 0.05\% & 1.67\% & 1.06\% & 1.42\% & 1.31\% & 1.05\% & 1.54\% & 1.61\% &  1.64\% \\
      & P90ob & & 72.17\% & -35.67\% & -26.22\% & -15.05\% & 10.41\% & -12.68\% & -32.28\% & -34.92\% &  -35.40\%  \\
      & P90ub & & 1.65\% & 9.40\% & 8.46\% & 5.15\% & 4.65\% & 5.05\% & 9.52\% & 10.17\% & 9.74\% \\
    \bottomrule
  \end{tabular}
  }
  \caption{Ablation studies for multi-magnitude mixture of experts on dataset \texttt{D1}.}
  \label{table:ablation}
\end{table*}

\FloatBarrier

\subsection{\texttt{D2} Ablations} \label{appendix:regional}
In this section, we present detailed experiment results for dataset \texttt{D2}.
Table \ref{tab:regional} summarizes the overall quantile loss from multiple models listed in Appendix \ref{appendix:ablation}. Overall, V11 performs best on P50 and V19 performs the best on P90. In contrast to V11, V16 and V18, V19 improves over ``Super Fast'' category. Despite the fact that the fraction of ``Super Fast'' series is smaller than 0.01\%, the impact of super fast products to the quantile loss metrics are large due to its high demand (large magnitude). For example, the loss of sales of such high demand product can easily lead to large monetary loss and customer dissatisfaction.

For all models, the major improvement on P90 comes from higher level of forecast, indicated by worse P90ob and better P90ub. With more than 90\% zero series, the benchmark model (V9) tends to provide lower level of forecast due to jointly training of products with different velocity. With sparse series routing, we avoid such impact of zero series in training. As a result, the forecast for non-zero categories are now higher and have lower quantile loss.

\begin{figure}
 \begin{subfigure}{0.95\columnwidth}
      \includegraphics[width=\linewidth]{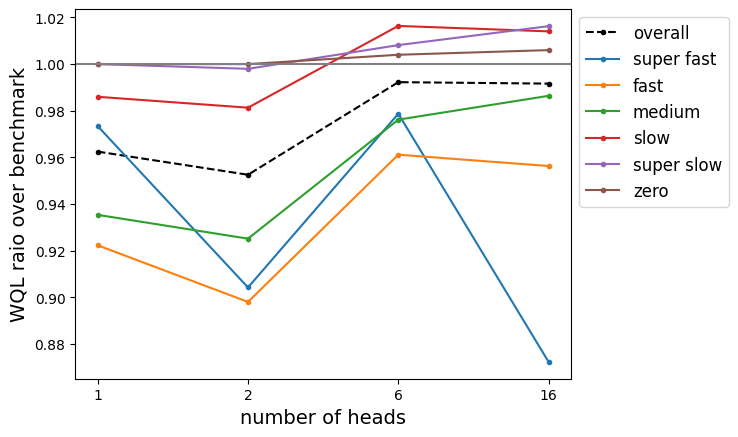}
      \caption{P50}
      \label{fig:reg_ablation_n_head_P50}
   \end{subfigure}
 \begin{subfigure}{0.95\columnwidth}
      \includegraphics[width=\linewidth]{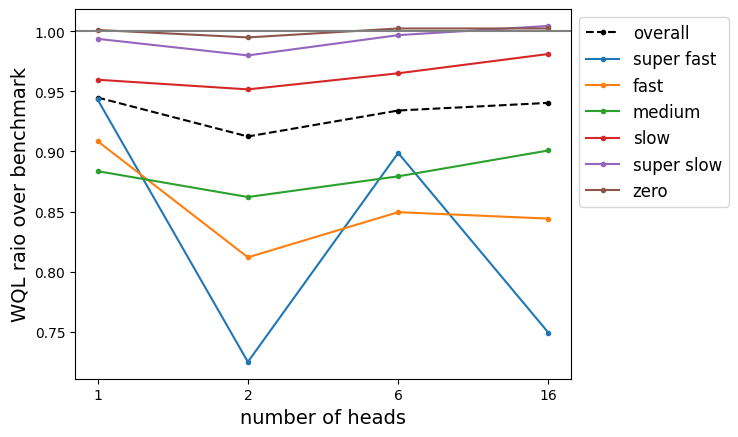}
      \caption{P90}
      \label{fig:reg_ablation_n_head_P90}
   \end{subfigure}
   \caption{WQL of all velocity groups over different number of heads in V19 model}
   \label{fig:reg_ablation_n_head}
\end{figure}

\begin{table}[H]
\centering
\begin{tabular}{r l ccccc}
    \toprule
    &            & \multicolumn{5}{c}{\bfseries Version} \\
    \addlinespace
    \bfseries Category & \bfseries Metric \textbf{}
      & \bfseries V9 & \bfseries V11 
      & \bfseries V16 & \bfseries V18 & \bfseries V19 \\
    \midrule
    \multirow{6}{*}{All}
      & P50ql & & \textbf{-1.36\%} & 0.87\% & -0.45\% &  -0.77\%\\
      & P90ql & & -3.39\% & -1.18\% & -5.49\% &  \textbf{-6.58\%} \\
    \addlinespace
      & P50ob & & -0.33\% & 11.08\% & 45.75\% & 32.86\% \\
      & P50ub & & -1.57\% & -1.21\% & -9.87\% & -7.63\%  \\
      & P90ob & & 6.97\% & 6.83\% & 26.88\% & 14.58\% \\
      & P90ub & & -6.37\% & -3.49\% & -14.82\% & -12.68\%  \\
    \cmidrule{2-7}
    \multirow{6}{*}{Super Fast}
      & P50ql & & 3.3\% & 14.2\% & 16.6\% & \textbf{-2.0\%} \\
      & P90ql & & -0.6\% & 7.9\% & 11.5\% & \textbf{-10.3\%} \\
    \addlinespace
      & P50ob & & -5.4\% & 28.7\% & -8.8\% & 6.3\%  \\
      & P50ub & & 5.1\% & 11.0\% & 22.0\% & -3.8\%\\
      & P90ob & & -8.8\% & 13.4\% & 9.2\% & 20\% \\
      & P90ub & & -0.3\% & 7.6\% & 11.5\% & -11.6\% \\
    \cmidrule{2-7}
    \multirow{6}{*}{Fast}
      & P50ql & & -2.9\% & 1.3\% & -0.8\% & \textbf{-3.9\%}  \\
      & P90ql & & -7.3\% & -3.3\% & -8.6\% & \textbf{-14.8\%} \\
    \addlinespace
      & P50ob & & -5.1\% & 1.2\% & 32.0\% & 34.1\% \\
      & P50ub & & -2.1\% & 1.4\% & -12.7\%  & -17.6\% \\
      & P90ob & & 6.6\% & 4.4\% & 31.6\% & 26.8\% \\
      & P90ub & & -9.8\% & -4.7\% & -15.9\% & -22.4\% \\
    \cmidrule{2-7}
    \multirow{6}{*}{Medium}
      & P50ql & & \textbf{-3.0\%} & 0.2\% & -2.7\% & -2.5\%  \\
      & P90ql & & -6.6\% & -4.4\% & \textbf{-13.0\%} & -12.2\% \\
    \addlinespace
      & P50ob & & -0.3\% & 11.1\% & 48.9\% & 32.5\% \\
      & P50ub & & -3.9\% & -3.4\% & -19.8\% & -14.0\% \\
      & P90ob & & 7.7\% & 10.1\% & 36.1\% & 22.3\% \\
      & P90ub & & -11.8\% & -9.7\% & -30.9\% & -24.7\% \\
    \cmidrule{2-7}
    \multirow{6}{*}{Slow}
      & P50ql & & \textbf{-0.5\%} & 0.8\% & 0.3 & 1.6\% \\
      & P90ql & & -1.8\% & 0.5\% & \textbf{-4.1\%} & -3.6\% \\
    \addlinespace
      & P50ob & & 2.9\% & 19.7\% & 60.7\% & 33.7\%\\
      & P50ub & & -1.0\% & -2.1\% & -9.0\% & -3.3\% \\
      & P90ob & & 4.1\% & 4.0\% & 11.0\% & 12.5\% \\
      & P90ub & & -4.9\% & -1.3\% & -3.9\% & -11.9\% \\
    \cmidrule{2-7}
    \multirow{6}{*}{Super Slow}
      & P50ql & & 0.2\% & 0.6\% & 0.3\% & 0.8\% \\
      & P90ql & & -0.7\% & 0.0\% & \textbf{-1.1\%} & -0.3\% \\
    \addlinespace
      & P50ob & & 11.0\% & 31.2\% & 41.0\% & 26.8\% \\
      & P50ub & & -0.1\% & -0.3\% & -1.0\% & 0.0\%\\
      & P90ob & & 5.6\% & 8.0\% & 11.0\%  & -1.4\%\\
      & P90ub & & -2.1\% & -1.8\% & -3.9\% & -0.1\%\\
    \cmidrule{2-7}
    \multirow{6}{*}{Zero}
      & P50ql & & 0.9\% & 0.5\% & 0.4\% & 0.4\% \\
      & P90ql & & 0.0\% & 0.1\% & \textbf{-0.3\%} & 0.2\% \\
    \addlinespace
      & P50ob & & 63.2\% & 41.2\% & 44.2\% & 30.1\% \\
      & P50ub & & 0.0\% & -0.1\% & -0.3\% & 0.0\% \\
      & P90ob & & 29.6\% & 18.6\% & 22.8\% & -2.9\% \\
      & P90ub & & -2.6\% & -1.6\% & -2.3\% & 0.5\% \\
    \bottomrule
  \end{tabular}
\caption{Ablation Studies for different methods dealing with sparse series on D2 dataset}
  \label{tab:regional}
\end{table}

All multi-head models (V16, V18 and V19) in Table \ref{tab:regional} have 6 heads. Here, we vary the number of heads for V19, the best performing model, and analyze the WQL in Figure \ref{fig:reg_ablation_n_head}. In general, we observe that WQL from 2 heads model is the lowest across almost all velocity groups in both P50 and P90. This suggests that for D2 dataset which has a large number of zero series, 2 heads are enough, and perform better than larger number of heads.

\end{document}